\documentclass[letterpaper, 10 pt, conference]{ieeeconf}  
\usepackage{graphicx}
\usepackage{booktabs} 
\usepackage{multirow} 
\usepackage{float}
\usepackage{siunitx}
\usepackage{amsmath}
\usepackage{placeins}
\usepackage{makecell}
\usepackage{array}
\usepackage{arydshln} 
\usepackage{tabularx} 

\usepackage{xcolor}

\IEEEoverridecommandlockouts                              %

\title{\LARGE \bf
Leveraging Tactile Sensing to Render both Haptic Feedback and Virtual Reality 3D Object Reconstruction in Robotic Telemanipulation}

\author{Gabriele Giudici$^{1}$, Aramis Augusto Bonzini$^{1}$, Claudio Coppola$^{2}$,\\ Kaspar Althoefer$^{1}$, Ildar Farkhatdinov$^{1,3}$, Lorenzo Jamone$^{1}$
\thanks{The study is funded by the UKRI EPSRC grants EP/R02572X/1 (NCNR), EP/V035304/1 (q-Arena) and Queen Mary University of London}
\thanks{$^1$ARQ (the Centre for Advanced Robotics @ Queen
Mary), School of Engineering and Materials Science, Queen Mary University of London, London, E14NS, UK (emails: \{g.giudici,a.a.bonzini,k.althoefer,i.farkhatdinov,l.jamone\}@qmul.ac.uk).}%
\thanks{$^2$ Humanoid AI. email: ccop@thehumanoid.ai}
\thanks{$^3$School of Biomedical Engineering and Imaging Sciences, King's College London, London, UK.}}

\begin{document}

\maketitle
\thispagestyle{empty}
\pagestyle{empty}

\begin{abstract}
Dexterous robotic manipulator teleoperation is widely used in many applications, either where it is convenient to keep the human inside the control loop, or to train advanced robot agents. So far, this technology has been used in combination with camera systems with remarkable success. On the other hand, only a limited number of studies have focused on leveraging haptic feedback from tactile sensors in contexts where camera-based systems fail, such as due to self-occlusions or poor light conditions like smoke. 
This study demonstrates the feasibility of precise pick-and-place teleoperation without cameras by leveraging tactile-based 3D object reconstruction in VR and providing haptic feedback to a blindfolded user. Our preliminary results show that integrating these technologies enables the successful completion of telemanipulation tasks previously dependent on cameras, paving the way for more complex future applications.
\end{abstract}

\section{INTRODUCTION}
Robotic teleoperation enables the remote control of a robotic system by a human operator for tasks that are dangerous, inaccessible, or intricate \cite{deng2021review}. This is required in numerous applications where current control techniques fall short in achieving autonomous robotic operations, where it is preferable to maintain a human in the loop, and where it is useful to transfer the human skills to the robot resorting to the learning by demonstration paradigm. 
Most teleoperation activities rely on vision-based systems, which, through cameras, allow manipulation experiments to be carried out at different levels of complexity \cite{darvish2023teleoperation}. 
Such systems are not infallible, and harsh visual conditions such as poor light, the presence of shadows, haze, smoke, radiation \cite{smith2020robotic} or even dirty or damaged camera lenses limit their reliability \cite{zang2019impact}. 
Tactile sensing can address some of these issues as it can provide measurements of properties that cannot be easily estimated with vision alone, such as, weight or texture information of the manipulated object surface \cite{luo2017robotic,li2020review}; for example, they can be used to reconstruct the 3D shape of an object \cite{Augusto_icdl}, \cite{Augusto_symmetries}, which could then be displayed using VR or AR technologies.
In addition, tactile sensing can be used to provide haptic and force feedback to the human operator, obtaining bilateral telerobotic systems capable of delivering an immersive haptic experience \cite{shahbazi2018systematic}, \cite{girbes2020haptic}.
\begin{figure} 
\centering

\includegraphics[width=0.80 \linewidth,keepaspectratio]{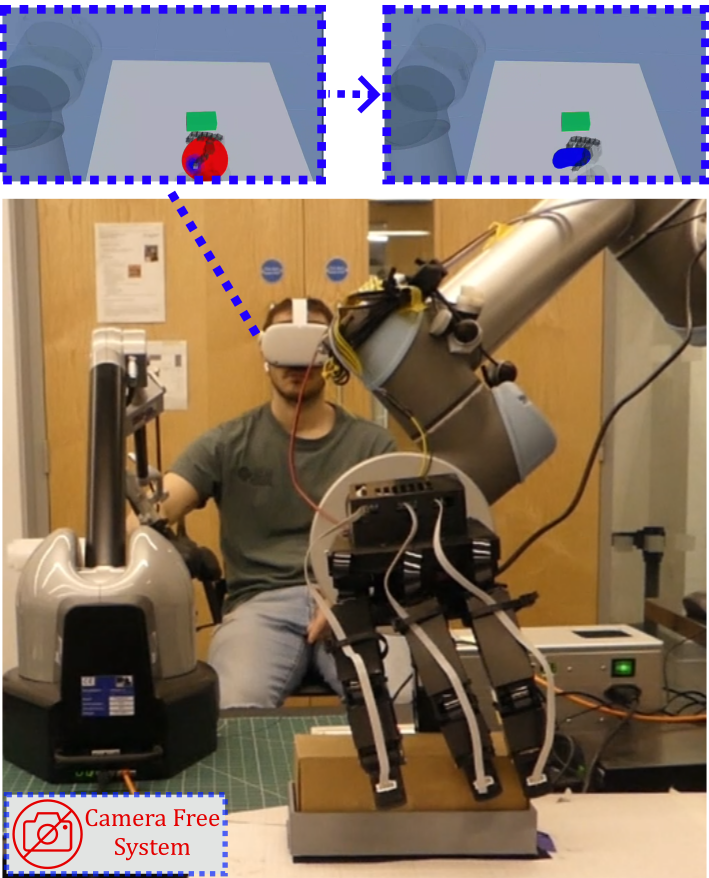} %
\caption{View of the human-operator setup (haptic interface with the glove and a VR visor) and the teleoperated robotic hand.
}
\label{operator_setup}
\end{figure}

So far, to the best of our knowledge, no other work has demonstrated the feasibility of conducting dexterous telemanipulation experiments in real-world settings without the aid of cameras or human vision of the target's real workspace.
Therefore, to address the current research gap and the significant demands posed by teleoperation in complete blindness, we have integrated a visual Virtual Reality (VR) visor into our established telerobotics setup \cite{giudici2023feeling}. The adopted bilateral teleoperation system exclusively relies on tactile sensing from the robot hand, omitting visual sensors. This system delivers crucial haptic feedback to the human user and facilitates accurate object manipulation. Moreover, we have developed a minimalist virtual environment to visualize the robotic arm in real-time during 3D object reconstruction and manipulation tasks. 
Utilizing only tactile sensors and Gaussian Process (GP), our system streams real-time partial 3D reconstructions of the shape of the manipulated object onto the visor while rendering haptic experience of the manipulated object, enhancing immersion and reducing fatigue.
This aim to mitigate the challenges associated with blind teleoperation, where it is challenging, if not impossible, for the teleoperator to manipulate objects accurately without haptic feedback alongside VR support.
In this work, we provide experimental evidence that a human operator, with prior experience in using vision-supervised teleoperation, can perform blind pick-and-place tasks accurately supported exclusively by haptic feedback and the VR visual representation. 

To summarize, the main contributions of this work are:
\begin{itemize}

    \item Integration of Virtual Reality into the teleoperation setup from \cite{giudici2023feeling}, so that tactile sensing on the robot hand is used not only to render haptic feedback through an exoskeletal glove but also to generate a real-time 3D reconstruction of the object shape via a Meta Quest 2 headset, thus providing rich feedback to the operator without cameras.
    
    \item A study demonstrating the feasibility of precise pick-and-place tasks using blind teleoperation, where the operator is isolated from the real scene and relies solely on tactile-based virtual object reconstruction and haptic feedback.

\end{itemize}
The paper is organized as follows: Section \ref{sec_RelWorks} reviews related work. Section \ref{sec_Methodology} outlines the telemanipulation system and VR elements. Section \ref{sec_Experiments} describes the experimental setup, and Section \ref{sec_EXPERIMENTAL_RESULTS} details the results. In Section \ref{sec_Discussion}, we discuss the findings, and Section \ref{sec_Conclusion} suggests future research directions.

\section{RELATED WORK} \label{sec_RelWorks}
The system we propose is pioneering, as no real-world robotic telemanipulation system in the literature,  to the best of our knowledge, avoids reliance on visual sensing.
However, in this section, we briefly introduce related works in two key areas:
autonomous robotic manipulation via tactile feedback alone (excluding visual cues) and multimodal telemanipulation.
We considered it unnecessary to compare the performance and accuracy of our work with studies employing vision-based approaches, as our work is not intended to be an alternative but rather a complement to systems utilizing cameras.

\subsection{Tactile Manipulation}
Although the interest in object manipulation is widely studied in the field of robotics \cite{mason2018toward}, \cite{billard2019trends}, \cite{suomalainen2022survey}, the exploration of manipulating objects without reliance on a vision system remains a relatively less explored field.
In \cite{murali2018learning}, the authors proposed a learning method based on a touch scan through which it was possible to identify the position of an object to grasp it stably. This work \cite{wu2019mat} introduced a tactile closed-loop method using deep reinforcement learning to improve grasping stability and overcome the limitations of vision-based robotic grasping systems prone to calibration errors and occlusion.
In this study \cite{pai2023tactofind}, the authors presented a robotic system able to localize and grasp freely moving objects only using only information gathered via local touch sensors without any visual feedback. In \cite{mao2024dexskills}, a framework to segment human demonstrations is presented to support the learning of primitive skills that compose long-horizon tasks. Although this framework does not require the use of cameras, 
during the telemanipulation demonstrations the operator has a direct vision of the workspace.
Different works have furthermore proven how it is possible through tactile information to reconstruct the 3D shape of an object \cite{luo2017robotic}. Amongst the various approaches \cite{yi2016active}, \cite{meier2011probabilistic}, \cite{Augusto_symmetries} which by using GP and point clouds were able to reconstruct the object with significant results. 
However, none of these methods have been integrated into a telerobotic setup that allows the human operator to manually control the robot and perceive a haptic experience of the manipulated object.

\subsection{Multimodal Telemanipulation}

The significance of a haptic experience made possible through touch is thoroughly explored in this review \cite{culbertson2018haptics}, and a wide variety of solutions for bilateral telerobotics have been proposed, as outlined in these reviews \cite{deng2021review, nahri2022review}. Various works have made use of tactile sensing to provide the operator with haptic feedback, such as in this work \cite{coppola2022affordable}, where haptic feedback was shown to reduce the physical effort of the operator during the grasping of small objects and in-hand manipulation tasks.
A prime example of progress in this field is the \$10 million ANA Avatar XPRIZE challenged global teams to develop avatar systems with high levels of manipulation abilities across ten challenging tasks. Notable examples include \cite{ParkXprize}, which utilized haptic feedback alongside visual input, the AvaTRINA Nursebot \cite{marques2023commodityXPrize}, which combined haptics with augmented reality, and the winning NimbRo Avatar system \cite{schwarz2023robust}, which excelled in telepresence.
Furthermore, multiple studies have been focusing on VR-based telemanipulation. Notably, \cite{whitney2018ros, lee2018implementation, buki_vr_TFDC} conducted proficiency remote telerobotics using cameras mounted on the robotic arm. Moreover, a multimodal study \cite{triantafyllidis2020study} proposed a large series of comparison experiments to evaluate the impact of combining audiovisual and haptic feedback on human telemanipulation performance in a virtual environment. The results showed that visual feedback via a VR visor had the most positive impact on operator performance. They also demonstrated that adding another feedback modality significantly benefits operator performance compared to using visual feedback alone.
However, these methods either rely partially on camera systems or have only been validated in simulation. To date, no research has demonstrated the feasibility of a human operator teleoperating a robot without any real-time visual feedback of the actual workspace. 

\section{METHODOLOGY} \label{sec_Methodology}
\subsection{Telemanipulation system} \label{sec3a}
\begin{figure}[t]
\centering
\includegraphics[width=1.0\linewidth,keepaspectratio]{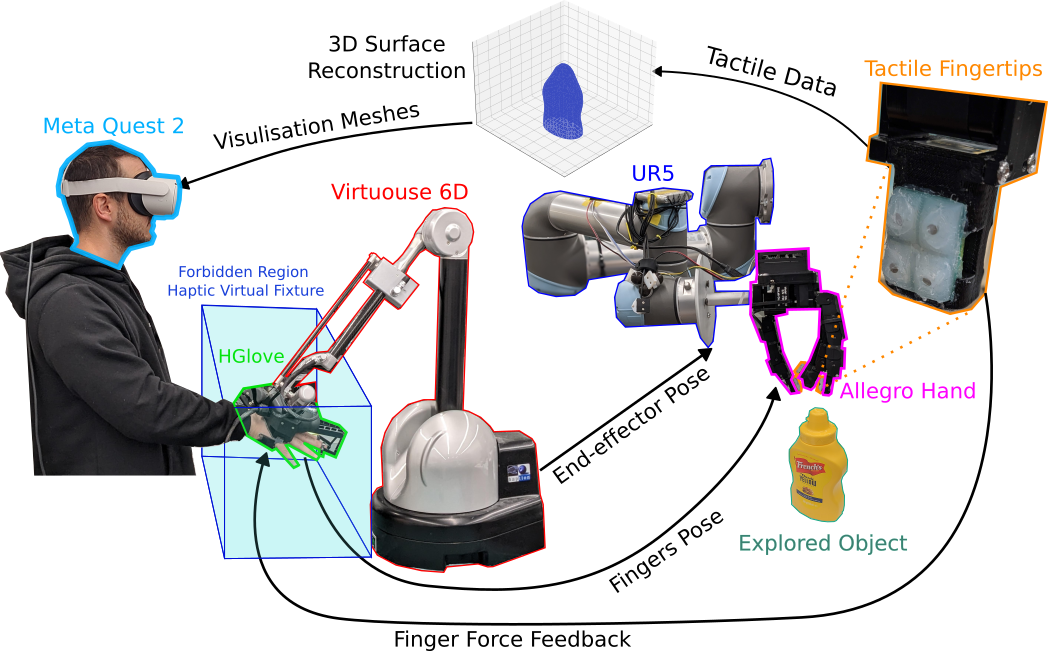}
\caption{
Control Scheme for Blind Bilateral Teleoperation
{: 
The operator controls the UR5 end effector using Virtuose 6D for pose and HGlove for Allegro Hand fingers. A haptic virtual fixture restricts movement beyond the blue box. Custom tactile sensors on the robotic hand provide kinesthetic feedback and real-time shape reconstruction. Visualization meshes streamed via Meta Quest 2 VR visor improve teleoperation transparency.
}}
\label{Controlmodel}
\end{figure}

In our experimental study, we used an enhanced version of the robotic setup illustrated in Fig.~\ref{Controlmodel}, previously validated in \cite{giudici2023feeling}.
The Leader system is composed by a Virtuose 6D \cite{Virtuose} and an exoskeletal glove HGlove \cite{HGlove} mechanically attached together. The Follower system comprises an Allegro robotic hand equipped with custom tactile sensors, attached to a UR5 robotic arm.
The tactile sensors employ the magnetic technology introduced in \cite{paulino17tactile,uskin}, and they are described in more details in \cite{giudici2023feeling}.
The mean normal force estimated by the tactile sensors measurement is rendered in the form of kinesthetic haptic feedback to the operator's fingers via HGlove. Virtuose 6D has the role of compensating for the weight of the glove and reduce the effort of the operator in keeping the arm in a non-neutral position during teleoperation. 
It also translates the user's hand movements into control commands for the UR5 arm in Cartesian space and creates a virtual haptic fixture to prevent movements in restricted areas.
Notably, the application of the RelaxedIk \cite{RelaxedIK} framework, based on a neural network, ensures the avoidance of self-collisions while computing the inverse kinematics essential for controlling the robot.
 An overview of the robotic setup employed is shown in Fig. \ref{Controlmodel}, differing from \cite{giudici2023feeling} by integrating the Meta Quest 2 visor with real-time 3D object reconstruction from haptic feedback in a virtual environment.

\subsection{Gaussian Process Shape Estimation} \label{sec3b}

The magnetic tactile sensors also allow for accurate contact detection.
When contact occurs, the locations of these contacts are obtained by calculating the Forward Kinematics (FK) of the arm-hand system.

Using this contact location data, the shape of the object can be reconstructed using Gaussian Process (GP) shape completion techniques.
This technique is a very commonly used method in the GP shape estimation literature (such as in selected works \cite{Augusto_icdl}, \cite{Augusto_symmetries}, \cite{martens2016geometric}, \cite{williams2006gaussian}).
With this method, all the $N$ contact points are registered into contact dataset $X$, and the object is estimated as a Signed Distance Field (SDF) where values of $Y=0$ correspond to the points on the surface of the object (i.e. where the contacts are observed).
Then, estimation of the object's surface is calculated as:

\begin{gather}
    \label{eq:mean}
    \mu(X_*) = m(X_*) + K_* ( K + \sigma_n^2I)^{-1} (Y-m(X)) \\
    \label{eq:var}
    \sigma^2(X_*) = K_{**} - K_*^{T}(K+ \sigma_n^2I)^{-1}K_{*}
\end{gather}

With $K$, $K_*$ and $K_{**}$ as the GP correlation matrices, $m(\cdot)$ as a surface estimation prior function, and and $\sigma_n^2$ the estimated measurement noise \cite{williams2006gaussianBook}.
For the purpose of this task, spherical priors, as first proposed by \cite{martens2016geometric}, were used to allow representation of the SDF Implicit Surface without the need of inserting artificial points inside or outside the surface under exploration.
The covariance function used to calculate the $K$ matrices is the 3-dimensional Thin-Plate Spline (TPS) model, commonly used in robotic shape reconstruction as shown in \cite{williams2006gaussian}.

The resulting $\mu(X_*)$ (Eq. \ref{eq:mean}) is used alongside a marching cubes algorithm to draw the estimated object on the screen.
The estimated surface variance $\sigma^2(X_*)$ is used to represent shape estimation uncertainty as the color of the resulting surface.
This is seen in Fig. \ref{vr_pov}, using blue and red colors to represent low and high surface variance zones respectively.

\subsection{Virtual Reality Setup}\label{sec3c}
To enhance teleoperation transparency, which refers to the degree of immersion and perception of the follower device as an extension of the operator's limbs, in scenarios lacking visual cues, we incorporated a Meta Quest 2 visor into the robotic setup.
This visor assumed a pivotal role by enabling the streaming of a virtual representation,  created within the environment of RViz, a 3D visualization tool for the Robot Operating System (ROS). As depicted in Fig.~\ref{vr_pov}, a viewpoint was chosen, guided by empirical considerations, to closely align the position of the robotic arm closely with that of the human operator's arm during teleoperation.
This virtual environment allows for the incorporation of various digital elements, such as timers and target markers, seamlessly integrated into the scene.
The visual setting designed for this research features a semi-transparent rendering of the robotic arm to minimize visual obstructions. Additionally, a red semi-sphere on the grey worktable serves as the Gaussian Process (GP) prior. This semi-sphere dynamically transforms to blue after contact measurements to approximate the manipulated object's shape. Two countdown timers in green and blue, accompanied by a green target marker resembling the object's shape, are also included in the scene.
This comprehensive augmentation collectively contributed to enhances the sense of transparency and control within the teleoperation framework, thereby mitigating the disorienting effects associated with tele-operating under blind constraints.

\section{EXPERIMENTS} \label{sec_Experiments}
\subsection{Experimental Setting}
The series of experiments conducted in this study focuses on the teleoperation of a real robot Follower under the constraint of a blind modality, meaning the absence of any camera system or real-world visual input. These experiments involved a participant who had prior proficiency in operating the system under visual supervision but had no previous training for the blind scenario.
Regarding the experimental configuration, the operator is seated at a station situated \SI{1.0}{m} from the base of the Leader apparatus, which encompasses the HGlove exoskeleton glove and the Virtuose 6D device. After the calibration procedure, which maps the movements of the operator's fingers with those of the Allegro Hand as described in \cite{giudici2023feeling}, the operator donned a Meta Quest 2 VR visor, as shown in Fig.~\ref{operator_setup}. The VR visor provided the operator with a visual perspective of the robotic arm and hand, presented within a simulated Rviz environment.
For the conduct of the experiments, an assistant had to supervise the setup in order to intervene in the event of a malfunction. In addition, the assistant was responsible for positioning the object in its starting position and replacing it at the end of each trial session.

\begin{figure*}[ht] 
\centering
\includegraphics[width=1\textwidth,keepaspectratio]{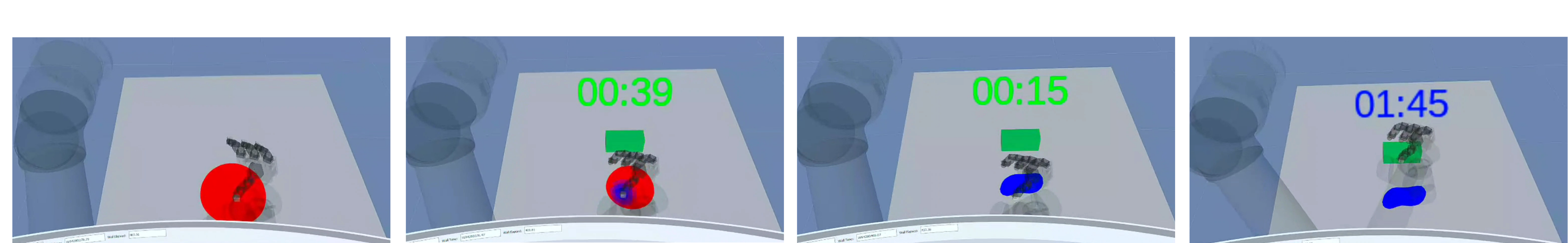} %
\caption{
Virtual Reality Scene: This series of images captures various stages of an experiment from the operator's viewpoint, recorded using the Meta Quest 2. On the grey worktable, a red semi-sphere renders the GP prior, which dynamically shifts to blue following contact measurements to approximate the shape of the manipulated object. The timer depicted is illustrative and does not correspond to an actual experiment.
}
\label{vr_pov}
\end{figure*}

Within the virtual realm shown in Fig.~\ref{vr_pov}, a fixed observation point was established to facilitate depth perception. A plane, coinciding with the work table used for the experiments, was displayed to provide a working spatial reference. A green marker with the shape and size of the manipulated object was placed on the virtual plane to provide a reference for the goal of the placement task. In addition, an initially red semi-sphere denotes a spatial region within where an object was located. 
The semi-sphere initially represents the GP Prior, which undergoes a blue colour change upon contact identification, rendering an approximation of the object surface generated by the GP. This colour change denotes an updated representation of the spatial feature of the object, estimated by the GP.
\subsection{Manipulated Objects}
In this preliminary study, we chose a lightweight, rigid rectangular cardboard object (\SI{0.30}{kg}) with minimal geometric complexity to ensure stable grasping and minimize potential damage to the robotic arm during experiments.
The object was presented in three configurations (Fig.~\ref{FIG: session}): leaning on its lateral face ($O1$, $15.5 \times 5.5$ cm), positioned on its larger base ($O2$, $15.5 \times 8.25$ cm), and on its smaller base ($O3$, $15.5 \times 8.25$ cm) in a vertical orientation.

The target marker was aligned with the object at a distance of four times the object's width in both $O1$ and $O2$ configurations, precisely \SI{22.0}{cm} from the object's center. To prevent undesired sliding, the object was placed on a base with an edge height of \SI{2.0}{cm} for $O1$ and $O2$, and \SI{3.0}{cm} for the vertical $O3$ configuration, as shown in Fig.~\ref{operator_setup}.

\subsection{Blind Teleoperated Pick and Place}

Prolonged use of virtual reality and the exoskeleton device can impose significant physical and cognitive demands on the operator, potentially compromising experimental performance. To mitigate these effects, daily sessions were restricted to a maximum duration of three hours, with individual sessions not exceeding one hour. Additionally, to prevent overtraining, two sessions were conducted on two days, separated by a one-month interval.

The experimental protocol consisted of three distinct sessions, each starting two hours following the conclusion of the previous one. 
The procedure was conducted over two separate days, $Day 1$ and $Day 2$, totalling six experimental sessions. In each session, the user encountered the three object configurations in different successions. For each configuration, the user was asked to perform the task five consecutive times. After completing the trials for a configuration, an assistant replaced the object and set up the next configuration.
Each experimental trial consisted of three consecutive phases. The first phase involved exploring the virtual environment to locate the object within a red semi-spherical space.
 Upon initial contact, a green countdown timer of $[00'20"]$ was activated and displayed in the virtual scene, marking the second phase. During this phase, the object's colour changed to reflect shape reconstructions updated via GP techniques based on new contact information.
Upon the expiration of the green countdown timer, it was succeeded by a second timer set to $[02'00"]$, this time in blue, signifying the commencement of the third and final phase.
After the green timer expired, a blue timer of $[02'00"]$ began, indicating the third and final phase, where the user aimed to pick up the object and align it with a designated green virtual marker shaped like the object.

This experimental design was conceived to comprehensively assess the user's performance in differentiating and manipulating the object in various configurations within the camera-blind teleoperation context. The structured phases and deliberate repetitions enabled a comprehensive evaluation of the user's adaptability and proficiency under the prescribed conditions. 

\section{EXPERIMENTAL RESULTS} 

\begin{figure*} 
\centering
\includegraphics[width=0.85\textwidth,keepaspectratio]{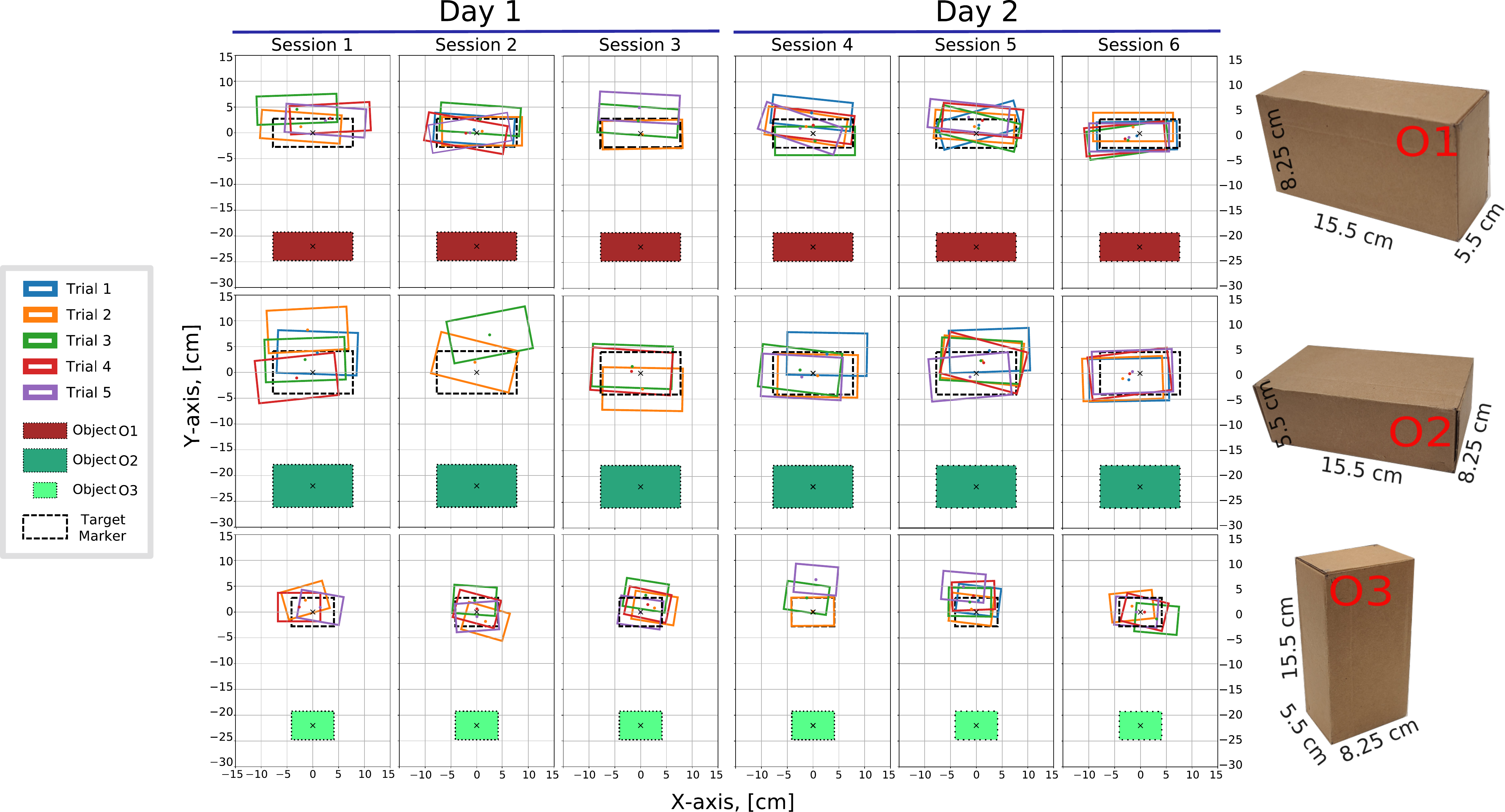} %
\caption{
Pick and Place Results: Each object is distinctly coloured, and the virtual representation of the target's base is depicted with a black dashed line. The contours of the object, placed after each trial, are illustrated using five different colours.
Failure of the task, meaning the object is not placed in its original O1, O2, or O3 configuration at the end of the trials, results in an omission of the contour representation on the graph. 
}
\label{FIG: session}
\end{figure*}

\begin{figure*} 
\centering
\includegraphics[width=0.9\textwidth,keepaspectratio]{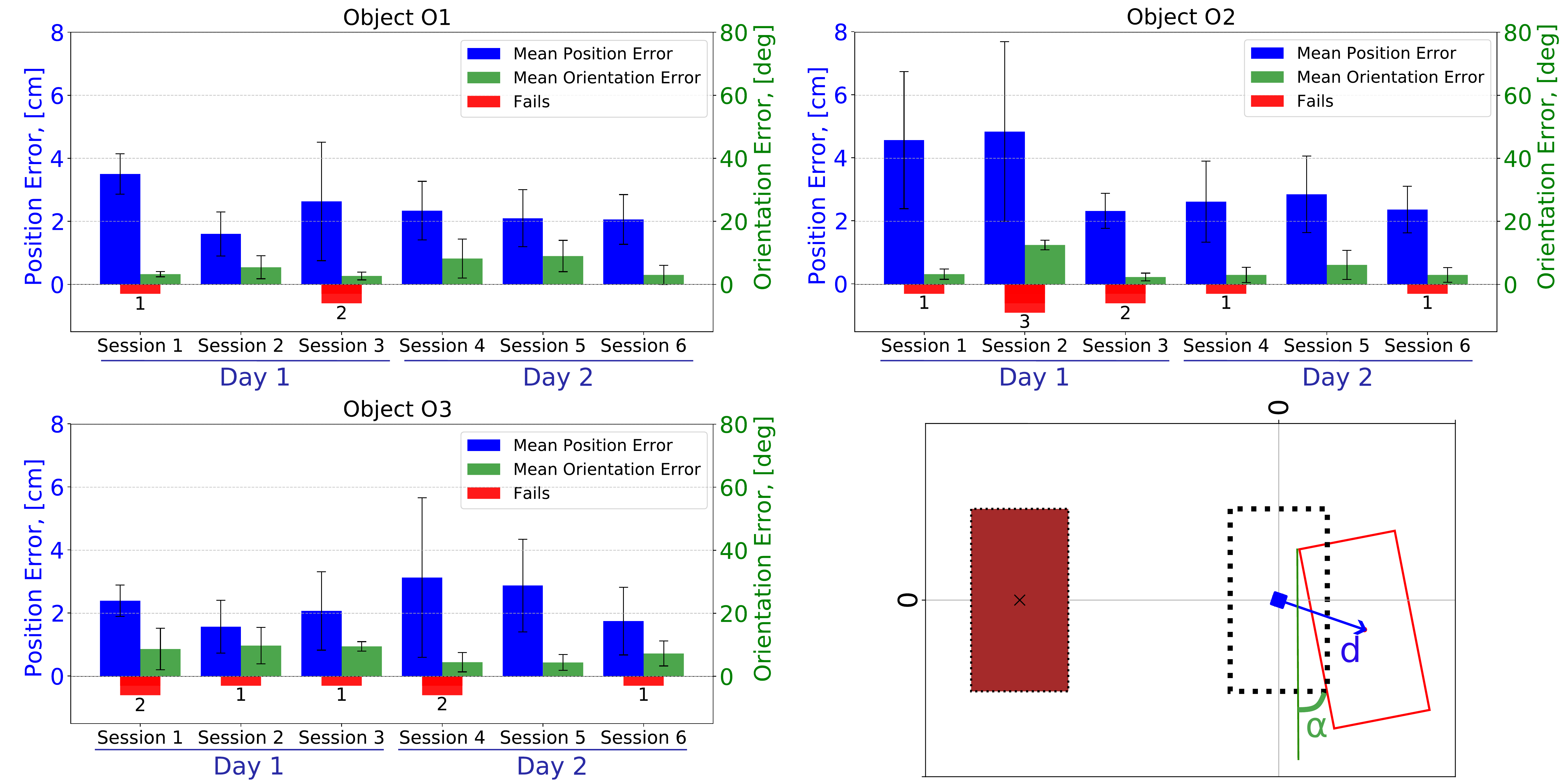} 

\caption{
Pick and Place Errors. Each of the three bar charts displays, for a specific object, the position error (d) in blue (mean and standard deviation of the five repetitions), the orientation error ($\alpha$) in green (mean and standard deviation of the five repetitions), and the number of failed attempts in red, for each experimental Session. The image on the bottom right corner explains visually our definition of position error (d) and orientation error ($\alpha$). 
}
\label{mean_error}
\end{figure*}

\label{sec_EXPERIMENTAL_RESULTS}

\begin{table}[ht]
\centering

\label{tab_combined}
\setlength{\tabcolsep}{2pt} 
\renewcommand{\arraystretch}{1.3} 
\fontsize{7}{7}\selectfont 

\begin{tabularx}{\columnwidth}{l|X X X|X X X|X X X}
\toprule
Metric & \multicolumn{3}{c|}{Object O1} & \multicolumn{3}{c|}{Object O2} & \multicolumn{3}{c}{Object O3} \\
\midrule
 & Day 1 & Day 2 & Overall & Day 1 & Day 2 & Overall & Day 1 & Day 2 & Overall \\
\midrule
\makecell[l]{Position\\Mean [cm]} & 2.49 & 2.16 & 2.31 & 3.88 & 2.63 & 3.15 & 1.98 & 2.56 & 2.28 \\
\hdashline
\makecell[l]{Position\\Std Dev [cm]} & 1.37 & 0.88 & 1.14 & 2.28 & 1.13 & 1.81 & 0.99 & 1.79 & 1.49 \\
\hdashline
\makecell[l]{Orientation\\Mean [deg]} & 4.01 & 6.73 & 5.52 & 5.0 & 4.23 & 4.55 & 9.36 & 5.38 & 7.28 \\
\hdashline
\makecell[l]{Orientation\\Std Dev [deg]} & 2.76 & 5.58 & 4.74 & 4.29 & 3.77 & 4.01 & 4.99 & 3.46 & 4.71 \\
\midrule
\makecell[l]{Completion\\Time Mean [s]} & 16.91 & 14.73 & 15.70 & 18.00 & 17.46 & 17.68 & 19.09 & 18.05 & 18.78 \\
\hdashline
\makecell[l]{Completion\\Time Std Dev} & 4.81 & 1.94 & 3.61 & 5.40 & 3.99 & 4.50 & 3.20 & 2.54 & 2.82 \\
\midrule
Failures & 3/15 & 0/15 & 3/30 & 6/15 & 2/15 & 8/30 & 4/15 & 3/15 & 7/30 \\
\bottomrule
\end{tabularx}
\caption{ Summary of performance metrics across O1, O2, and O3}
\end{table}
\vspace{-6mm}

The objective this experimental work is to demonstrate the feasibility of conducting blind pick-and-place teleoperation using a real robot, relying solely on haptic feedback without cameras. 
We visually present in Fig. \ref{FIG: session} the outcomes of our experiments for each object, grouped in rows, across multiple sessions grouped in columns. A total of 15 trials per object were conducted daily, resulting in 30 trials per object and a total of 90 trials across all objects.
In the bar graphs presented in Fig. \ref{mean_error}, we present the mean placement position and orientation errors concerning the target for the successful tasks, accompanied by the count of failures recorded for each session.
Furthermore, in light of the distinct outcomes observed for each object, a summary of performance metrics across objects O1, O2, and O3, including daily error rates, mean positions and orientations errors, overall completion times, and failure rates have been documented in Table~\ref{tab_combined}.

In our analysis, we conducted a series of two-way Analysis Of Variance (ANOVA) with object type (Object) and experimental day (Day) as factors, analyzing Position Error, Orientation Error, Failures, and Time as dependent variables (see Table \ref{tab:anova_results}). ANOVA results for Position Error and Orientation Error showed no statistically significant effects of object type or day, nor significant interactions, indicating consistent accuracy across object types and days. However, a borderline significant interaction for Orientation Error ($p = 0.040$) suggests a minor influence of object type on orientation errors across days. Regarding Failures, the day factor showed a significant effect ($p = 0.035$), indicating variation in failure rates between days. The analysis of Time revealed a significant effect of object type ($p = 0.016$), indicating that different objects influenced task completion time.

\begin{table}[h]
\centering

\begin{tabular}{lcccc}
\toprule
\textbf{Variable} & \textbf{Sum of Squares} & \textbf{df} & \textbf{F-value} & \textbf{p-value} \\
\midrule
Position Error& & & & \\
C(Object) & 1143.75 & 2 & 2.549 & 0.086 \\
C(Day) & 168.21 & 1 & 0.750 & 0.390 \\
C(Object):C(Day) & 930.05 & 2 & 2.073 & 0.134 \\
\midrule
Orientation Error & & & & \\
C(Object) & 85.42 & 2 & 2.104 & 0.130 \\
C(Day) & 4.78 & 1 & 0.235 & 0.629 \\
C(Object):C(Day) & 137.08 & 2 & 3.376 & 0.040 \\
\midrule
Completion Time & & & & \\
C(Object) & 120.59 & 2 & 4.387 & 0.016 \\
C(Day) & 24.43 & 1 & 1.778 & 0.187 \\
C(Object):C(Day) & 10.89 & 2 & 0.396 & 0.674 \\
\midrule
Failures & & & & \\
C(Object) & 0.47 & 2 & 1.500 & 0.229 \\
C(Day) & 0.71 & 1 & 4.571 & 0.035 \\
C(Object):C(Day) & 0.16 & 2 & 0.500 & 0.608 \\
\bottomrule
\end{tabular}
\caption{ ANOVA results for Position Error, Orientation Error, Completion Time and Failures.}
\label{tab:anova_results}
\end{table}

\vspace{-8mm}
\section{DISCUSSION} \label{sec_Discussion}
We conducted 90 trials over two daily sessions separated by a one-month interval to assess long-term learning trends — primarily observed as a decrease in failed attempts — and to mitigate operator bias from cognitive and environmental conditions. The trials involved the rectangular object in three distinct configurations. Overall, our approach achieved an 80\% success rate, with 18 failures out of 90 trials. Additionally, an improvement in execution performance was observed between the first daily session, which had a success rate of 71.11\%, and the second daily session, which achieved a higher success rate of 88.89\%.
By cross-referencing the information from the bar graph shown in Fig. \ref{mean_error} and ANOVA analysis, it is possible to assert that:
\begin{itemize}
    \item 
    There is no clear indication of a learning curve in positioning and orientation errors within the same experimental day. The exact impact of fatigue on performance and task completion remains difficult to determine.
    \item 
    Conversely, a significant decrease in the number of failed attempts indicates an improvement trend in manipulating objects O1 and O2.
    \item 
    A significant improvement in orientation accuracy is observed for object O3, with the average error decreasing from $9.63^\circ$ to $5.39^\circ$. This improvement can be attributed to a more comprehension of the object-task relationship.
   \item 
   Experimental findings suggest that objects O1 and O2 are easier to manipulate due to their larger bases offering greater stability during placement tasks. This is supported by improvements observed on the second experimental day (sessions 4–6), characterized by reduced mean errors and increased consistency, as indicated by lower standard deviations.
  \item 
  Object O3 presents a significantly greater challenge due to its tighter dimensions, necessitating higher dexterity for precise grasping and enhanced accuracy during placement on its base. Although the number of failed trials has marginally decreased and the orientation error of correctly placed instances has reduced, no discernible overall improvement is observed for Object O3.
 \item 
 The completion time for the pick-and-place task shows minimal variation across test scenarios and object configurations, remaining between \SI{15}{s} and \SI{20}{s} for all successful trials, as indicated in Table \ref{tab_combined}.
\end{itemize}

In conclusion, the experimental results indicate that the geometric characteristics of the objects significantly influence the success rate of pick-and-placeexperiments. Specifically, for Object O1, the mean position error was \SI{2.31}{cm} with a mean orientation error of \SI{5.52}{deg}; Object O2 had a mean position error of \SI{3.15}{cm} and a mean orientation error of \SI{4.55}{deg}; and Object O3 showed a mean position error of \SI{2.28}{cm} with an orientation error of \SI{7.28}{deg}. Overall, across all objects, the average position and orientation errors were below \SI{3}{cm} and \SI{6}{deg}, respectively.

\section{CONCLUSIONS} \label{sec_Conclusion}
The capability to teleoperate a robot remotely using only haptic feedback, without visual input, is essential for addressing scenarios where vision-based solutions are prone to failure, such as low-light or low-visibility conditions (e.g., smoke, radiation, occlusions).
In this work, we have demonstrated that integrating different teleoperation technologies enables performing simple pick-and-place tasks without relying on cameras. Specifically, we developed a telerobotic setup that includes a virtual fixture to ensure safety during unsupervised manipulation, a VR environment, and a model for reconstructing objects based on local haptic information.
This study lays the groundwork for future research to overcome current limitations of our integrated systems. 
In future work, we intend to develop a more immersive VR environment to improve three-dimensionality and depth perception, enabling more effective reconstruction of complex, moving objects, potentially by fully sensorizing the robotic hand.
Finally, we intend to conduct user studies with non-expert participants to evaluate system performance and address challenges related to higher precision and contact-intensive tasks like insertion.

\FloatBarrier

\bibliographystyle{IEEEtran}
\bibliography{blind_teleop}

\end{document}